\documentclass{sig-alternate-05-2015}
\usepackage{epstopdf}
\usepackage{array} 
\usepackage{color}
\usepackage{hyperref}
\usepackage{graphicx}
\usepackage[numbers, sort, comma, square]{natbib}
\usepackage{url}
\usepackage{amsmath}
\usepackage[inline]{enumitem}
\usepackage{framed}

\usepackage{cleveref}
\usepackage{xcolor}
\hypersetup{
	colorlinks,
	linkcolor={red!50!black},
	citecolor={blue!50!black},
	urlcolor={blue!80!black}
}

\newcommand{\RR}{\mathbb{R}}

\newcommand{\ue}{\tilde{u}}
\newcommand{\te}{\tilde{t}}
\newcommand{\ve}{\tilde{v}}
\newcommand{\ee}{\tilde{e}}

\makeatletter
\def\@copyrightspace{\relax}
\makeatother

\begin{document}


\clubpenalty=10000 
\widowpenalty = 10000

\title{{\ttlit Ask the GRU}: Multi-Task Learning for \\Deep Text Recommendations}
%
%
%
%
%

\numberofauthors{3} 
%
\author{
%
%
\alignauthor
Trapit Bansal \\
\email{tbansal@cs.umass.edu}
\alignauthor
David Belanger \\
\email{belanger@cs.umass.edu}
\alignauthor
Andrew McCallum \\
\email{mccallum@cs.umass.edu}
\and
	\affaddr{ College of Information and Computer Sciences, University of Massachusetts Amherst} \\	
}


\maketitle
\begin{abstract}
	
In a variety of application domains the content to be recommended to users is associated with text. This includes research papers, movies with associated plot summaries, news articles, blog posts, etc. Recommendation approaches based on latent factor models can be extended naturally to leverage text by employing an explicit mapping from text to factors. This enables recommendations for new, unseen content, and may generalize better, since the factors for all items are produced by a compactly-parametrized model. Previous work has used topic models or averages of word embeddings for this mapping.  In this paper we present a method leveraging deep recurrent neural networks to encode the text sequence into a latent vector, specifically gated recurrent units (GRUs) trained end-to-end on the collaborative filtering task. For the task of scientific paper recommendation, this yields models with significantly higher accuracy. In cold-start scenarios, we beat the previous state-of-the-art, all of which ignore word order. Performance is further improved by multi-task learning, where the text encoder network is trained  for a combination of content recommendation and item metadata prediction. This regularizes the collaborative filtering model, ameliorating the problem of sparsity of the observed rating matrix.

\end{abstract}

%
%
%



\section{Introduction}
Text recommendation is an important problem that has the potential to drive significant profits for e-businesses through increased user engagement. Examples of text recommendations include recommending blogs, social media posts \cite{guy2010social}, news articles \cite{phelan2009using, bansal2015content}, movies (based on plot summaries), products (based on reviews) \cite{mcauley2013hidden} and research papers \cite{wang2011collaborative}. 

Methods for recommending text items can be broadly classified into collaborative filtering (CF), content-based, and hybrid methods.
Collaborative filtering \cite{koren2009matrix} methods use the user-item rating matrix to construct user and item profiles from past ratings.
Classical examples of this include matrix factorization methods \cite{koren2009matrix,mnih2007probabilistic} which completely ignore text information and rely solely on the rating matrix.
Such methods suffer from the \textit{cold-start} problem -- how to rank unseen or unrated items -- which is ubiquitous in most domains.
Content-based methods \cite{balabanovic1997fab, mooney2000content}, on the other hand, use the item text or attributes, and make recommendations based on similarity between such attributes, ignoring data from other users.
Such methods can make recommendations for new items but are limited in their performance since they cannot employ similarity between user preferences \cite{basu1998recommendation, schein2002methods, wang2011collaborative}. Hybrid recommendation systems seek the best of both worlds, by leveraging both item content and user-item ratings \cite{basilico2004unifying, basu1998recommendation, wang2011collaborative, wang2015collaborative, melville2002content}. Hybrid recommendation methods that consume item text for recommendation often ignore word order \cite{melville2002content, wang2011collaborative, gopalan2014content, wang2015collaborative}, and either use bags-of-words as features for a linear model~\cite{melville2002content, agarwal2009regression} or define an unsupervised  learning objective on the text such as a topic model \cite{wang2011collaborative, gopalan2014content}. Such methods are unable to fully leverage the text content, being limited to bag-of-words sufficient statistics \cite{wallach2006topic}, and furthermore unsupervised learning is unlikely to focus on the aspects of text relevant for content recommendation. 

In this paper we present a method leveraging \textit{recurrent neural networks} (RNNs) \cite{werbos1990backpropagation} to represent text items for collaborative filtering.
In recent years, RNNs have provided substantial performance gains in a variety of natural language processing applications such as language modeling \cite{mikolov2010recurrent} and machine translation \cite{cho2014learning}. RNNs have a number of noteworthy characteristics: (1) they are sensitive to word order, (2) they do not require hand-engineered features, (3) it is easy to leverage large unlabeled datasets, by pretraining the RNN parameters with unsupervised language modeling objectives \cite{dai2015semi}, (4) RNN computation can be parallelized on a GPU, and (5) the RNN applies naturally in the cold-start scenario, as a feature extractor, whenever we have text associated with new items. 

Due to the extreme data sparsity  of content recommendation datasets~\cite{bell2007lessons}, regularization is also an important consideration. This is particularly important for deep models such as RNNs, since these high-capacity models are prone to overfitting. Existing hybrid methods have used unsupervised learning objectives on text content to regularize the parameters of the recommendation model~\cite{mcauley2013hidden, ling2014ratings,almahairi2015learning}. However, since we consume the text directly as an input for prediction, we can not use this approach. Instead, we provide regularization by performing multi-task learning combining collaborative filtering with a simple side task: predicting item meta-data such as genres or item tags. Here, the network producing vector representations for items directly from their text content is shared for both tag prediction and recommendation tasks. This allows us to make predictions in cold-start conditions, while providing regularization for the recommendation model.

We evaluate our recurrent neural network approach on the task of scientific paper recommendation using two publicly available datasets, where items are associated with text abstracts~\cite{wang2011collaborative, wang2015collaborative}. We find that the RNN-based models yield up to \textbf{34\%} relative-improvement in Recall@50 for cold-start recommendation over collaborative topic regression (CTR) approach of \citet{wang2011collaborative} and a word-embedding based model model \cite{weston2011wsabie}, while giving competitive performance for warm-start recommendation.
We also note that a simple linear model that represents documents using an average of word embeddings trained in a completely supervised fashion~\cite{weston2011wsabie}, obtains competitive results to CTR. Finally, we find that multi-task learning improves the performance of all of the models significantly, including the baselines.

\section{Background and Related Work}

\subsection{Problem Formulation and Notation}
This paper focuses on the task of recommending items associated with text content. The $j$-th text item is a {\it sequence} of $n_j$ word tokens, $X_j = (w_{1}, w_{2}, \ldots, w_{n_j})$ where each token is one of $V$ words from a vocabulary.
Additionally, the text items may be associated with multiple {\it tags} (user or author provided). If item $j \in [N_d]$ has tag $l \in [N_t]$ then we denote it by $t_{jl} = 1$ and $0$ otherwise. 

There are $N_u$ users who have liked/rated/saved some of the text items. The rating provided by user $i$ on item $j$ is denoted by $r_{ij}$. We consider the implicit feedback \cite{hu2008collaborative, rendle2009bpr} setting, where we only observe whether a person has viewed or liked an item and do not observe explicit ratings. $r_{ij}=1$ if user $i$ liked item $j$ and $0$ otherwise.
Denote the user-item matrix of likes by $R$.
Let $R^{+}_i$ denote the set of all items liked by user $i$ and $R^{-}_i$ denote the remaining items.

The recommendation problem is to find for each user $i$ a personalized ranking of all unrated items, $j \in R^{-}_i$, given the text of the items $\{X_j\}$, the matrix of users' previous likes $\{R_i\}$ and the tagging information of the items $\{t_{il}\}$.

The methods we consider will often represent users, items, tags and words by $K$-dimensional vectors $\ue_i$, $\ve_j$, $\te_l$ and $\ee_w$ $\in \RR^K$, respectively. We will refer to such vectors as~\textit{embeddings}. All vectors are treated as column vectors. $\sigma(.)$ will denote the sigmoid function, $\sigma(x) = \frac{1}{1 + e^{-x}}$.



\subsection{Latent Factor Models}
Latent factor models \cite{koren2009matrix} for content recommendation learn $K$ dimensional vector embeddings of items and users: 
\begin{align} \label{eq:pmf_r}
	\hat{r}_{ij} = b_i + b_j + \ue_i^T\ve_j,
\end{align}
$b_i, b_j$ are user and item specific biases, and $\ue_i$ is the vector embedding for user $i$ and $\ve_j$ is the embedding of item $j$. 

A simple method for learning the model parameters, $\pmb{\theta} = \{b_i, b_j, \ue, \ve\}$, is to specify a cost function and perform stochastic gradient descent. 
For implicit feedback, an unobserved rating might indicate that either the user does not like the item or the user has never seen the item.
In such cases, it is common to use a \textit{weighted} regularized squared loss \cite{wang2011collaborative, hu2008collaborative}: 
\begin{align} \label{eq:cost_mf}
	C_R(\pmb{\theta}) = \frac{1}{|R|} \sum_{(i,j) \in R} c_{ij} (\hat{r}_{ij} - r_{ij})^2  + \Omega(\pmb{\theta})
\end{align}
Often, one uses $c_{ui} = a$ for observed items and $c_{ui} = b$ for unobserved items, with $b \ll a$ \cite{wang2011collaborative, wang2015collaborative}, signifying the uncertainity in the unobserved ratings.
$\Omega(\pmb{\theta})$ is a regularization on the parameters, for example in PMF \cite{mnih2007probabilistic}
the embeddings are assigned Gaussian priors, 
which leads to a $\ell_2$ regularization.
Some implicit feedback recommendation systems use a ranking-based loss instead \cite{rendle2009bpr, weston2011wsabie}.
The methods we propose can be trained with any differentiable cost function. We will use a weighted squared loss in our experiments to be consistent with the baselines \cite{wang2011collaborative}.

\subsection{The Cold Start Problem} \label{sec:cold_start}
In many applications, the factorization~\eqref{eq:pmf_r} is unusable, since it suffers from the \textit{cold-start} problem \cite{schein2002methods,melville2002content}:~new or unseen items can not be recommended to users because we do not have an associated embedding. This has lead to increased interest in hybrid CF methods which can leverage additional information, such as item content, to make cold-start recommendations. 
In some cases, we may also face a cold-start problem for new users. Though we do not consider this case, the techniques of this paper can be extended naturally to accommodate it whenever we have text content associated with users. We consider:
\begin{align} \label{eq:pmf_f}
	\hat{r}_{ij} = b_i + b_j + \ue_i^Tf(X_j),
\end{align}
Where $f(\cdot)$ is a vector-valued function of the item's text. For differentiable $f(\cdot)$,~\eqref{eq:pmf_f} can also be trained using~\eqref{eq:cost_mf}. Throughout the paper, we will refer to $f(\cdot)$ as an \textit{encoder}. 
Existing hybrid CF methods \cite{shi2014collaborative, agarwal2009regression, schein2002methods, rendle2010factorization} which use item metadata take this form. In such cases, $f(.)$ is a linear function of manually extracted item features. 
For example, \citet{agarwal2009regression, gantner2010learning} incorporate side information through a linear regression based formulation on metadata like category, user's age, location, etc.
\citet{rendle2010factorization} proposed a more general framework for incorporating higher order interactions among features in a factor model. Refer to \citet{shi2014collaborative}, and the references therein, for a recent review on such hybrid CF methods.

Our experiments compare to collaborative topic regression (CTR) \cite{wang2011collaborative}, a state-of-the-art technique that simultaneously factorizes the item-word count matrix (through probabilistic topic modeling) and the user-item rating matrix (through a latent factor model). 
By learning low-dimensional (topical) representations of items, CTR is able to provide recommendations to unseen items.

\subsection{Regularization via Multi-task Learning}
\label{sec:back-mtl}

Typical CF datasets are highly sparse, and thus it is important to leverage all available training signals~\cite{bell2007lessons}. In many applications, it is useful to perform multi-task learning~\cite{caruana1997multitask} that combines CF and auxiliary tasks, where a shared feature representation for items (or users) is used for all tasks. Collective matrix factorization \cite{singh2008relational} jointly factorizes multiple observation matrices with shared entities for relational learning. \citet{ma2008sorec} seek to predict side information associated with users. Finally, \citet{mcauley2013hidden} used topic models and \citet{almahairi2015learning} used language models on review text.

In many applications, text items are associated with tags, including research papers with keywords, news articles with user or editor provided labels, social media posts with hash-tags, movies with genres, etc. These can be used as features $X_j$ in~\eqref{eq:pmf_f} \cite{agarwal2009regression, rendle2010factorization}. However, there are considerable drawbacks to this approach. First, tags are often assigned by users, which may lead to a cold-start problem~\cite{krestel2009latent}, since new items have no annotation.  Moreover, tags can be noisy, especially if they are user-assigned, or too general \cite{bansal2015content}. 

While tag annotation may be unreliable and incomplete as input features, encouraging items' representations to be predictive of these tags can yield useful regularization for the CF problem. Besides providing regularization, this multi-task learning approach is especially useful in cold-start scenarios, since the tags are only used at train time and hence need not be available at test time. In Section~\ref{sec:mtl2} we employ this approach.

\subsection{Deep Learning}

In our work, we represent the item-to-embedding mapping $f(\cdot)$ using a deep neural network. See~\cite{Goodfellow-et-al-2016-Book} for a comprehensive overview of deep learning methods. 
We provide here a brief review of deep learning for recommendation systems.

Neural networks have received limited attention from the recommendation systems community.
\cite{salakhutdinov2007restricted} used restricted Boltzmann machines as one of the component models to tackle the Netflix challenge.
Recently, \cite{sedhain2015autorec, wu2016cdae} proposed denoising auto-encoder based models for collaborative filtering which are trained to denoise corrupted versions of entire sparse vectors of user-item likes or item-user likes (i.e.~rows or columns of the $R$ matrix). However, these models are unable to handle the cold-start problem.
\citet{wang2015collaborative} addresses this by incorporating a bag-of-words autoencoder in the model within a Bayesian framework.
\citet{elkahky2015multi} proposed to use neural networks on manually extracted user and item feature representations for content based multi-domain recommendation.
\citet{dziugaite2015neural} proposed to use a neural network to learn the similarity function between user and item latent factors.
\citet{van2013deep, wang2014improving} developed music recommender systems which use features extracted from the music audio using convolutional neural networks (CNN) or deep belief networks.
However, these methods process the user-item rating matrix in isolation from the content information and thus are unable to exploit the direct interaction between item content and ratings \cite{wang2015collaborative}.
\citet{weston2014tagspace} proposed a CNN based model to predict hashtags on social media posts and found the learned representations to also be useful for document recommendation.
Recently, \citet{HeMcA16} used image-features from a separately trained CNN to improve product recommendation and tackle cold-start.
\citet{almahairi2015learning} used neural network based language models \cite{mikolov2013distributed, mikolov2010recurrent} on review text to regularize the latent factors for product recommendation, as opposed to using topic models, as in \citet{mcauley2013hidden}.
They found that RNN based language models perform poorly as regularizers and word embedding models \citet{mikolov2013distributed} perform better.


%

%
%
%
%

\section{Deep Text Representation for\\ Collaborative Filtering}
This section presents neural network-based encoders for explicitly mapping an item's text content to a vector of latent factors. This allows us to perform cold-start prediction on new items. In addition, since the vector representations for items are tied together by a shared parametric model, we may be able to generalize better from limited data.



As is standard in deep learning approaches to NLP, our encoders first map input text $X_j = (w_{1}, w_{2}, \ldots, w_{n_j})$ to a sequence of $K_w$-dimensional embeddings \cite{mikolov2013distributed}, $(e_{1}, e_{2}, \ldots, e_{n_j})$, using a lookup table with one vector for every word in our vocabulary. Then, we define a transformation that collapses the sequence of embeddings to a single vector, $g(X_j)$. 

In all of our models, we maintain a separate item-specific embedding $\ve_j$, which helps capture user behavior that cannot be modeled by content alone \cite{wang2011collaborative}. Thus, we set the final document representation as:
\begin{align} \label{eq:embed_avg_shift}
	f(X_j) = g(X_j) + \ve_j
\end{align}
For cold-start prediction, there is no training data to estimate the item-specific embedding and we set $\ve_j = 0$ \cite{wang2011collaborative, wang2015collaborative}.

\subsection{Order-Insensitive Encoders} \label{sec:order_insensitive}
A simple order-insensitive encoder of the document text can be obtained by averaging word embeddings: 
\begin{align} \label{eq:embed_avg}
g(X_j) = \frac{1}{|X_j|} \sum_{w \in X_j} e_w.
\end{align}
This corresponds exactly to a linear model on a bag-of-words representation for the document. However, using the representation~\eqref{eq:embed_avg} is useful because the word embeddings can be pre-trained, in an unsupervised manner, on a large corpus \cite{collobert2011natural}.	
Note that~\eqref{eq:embed_avg} is similar to the embedding-based model used in \citet{weston2014tagspace} for hastag prediction. 

Note that CTR ~\cite{wang2011collaborative}, described in \ref{sec:cold_start}, also operates on bag-of-words sufficient statistics. Here, it does not have an explicit parametric encoder $g(\cdot)$ from text to a vector, but instead defines an implicit mapping via the process of doing posterior inference in the probabilistic topic model.

\subsection{Order-Sensitive Encoders}
 \label{sec:order_sensitive}

\begin{figure*}[t!]
	\includegraphics[width=\textwidth]{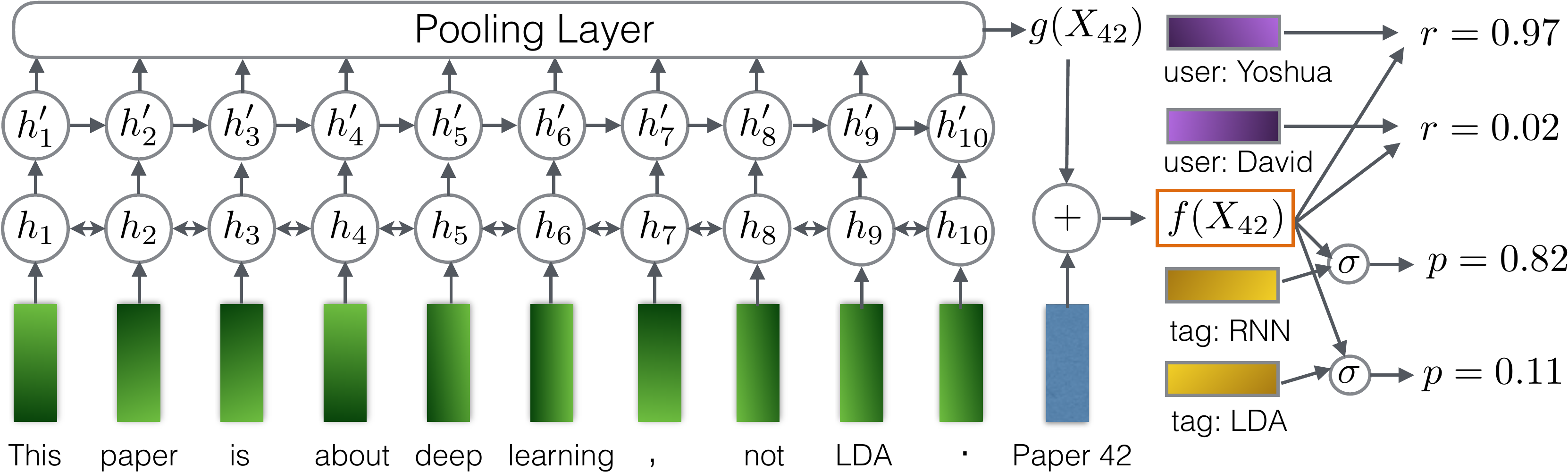}
	\caption{Proposed architecture for text item recommendation. Rectangular boxes represent embeddings. Two layers of RNN with GRU are used, where the first layer  is a bi-directional RNN. The output of all the hidden units at the second layer is pooled to produce a text encoding which is combined with an item-specific embedding to produce the final representation $f(X)$. Users and tags are also represented by embeddings, which are combined with the item representation to do tag predection and recommendation.}
	\label{fig:model}
\end{figure*}

Bag-of-words models are limited in their capacity, as they cannot distinguish between sentences that have similar unigram statistics but completely different meanings \cite{wallach2006topic}.
As a toy example, consider the research paper abstracts: ``This paper is about deep learning, not LDA" and ``This paper is about LDA, not deep learning''. They have the same unigram statistics but would be of interest to different sets of users. A more powerful model that can exploit the additional information inherent in word order would be expected to recognize this and thus perform better recommendation.  

In response, we parametrize $g(\cdot)$ as a recurrent neural network (RNN). It reads the text one word at a time and produces a single vector representation. RNNs can provide impressive compression of the salient properties of text. For example, accurate translation of an English sentence can be performed by conditioning on a single vector encoding~\cite{sutskever2014sequence}. 

The extracted item representation is combined with a user embedding, as in \eqref{eq:pmf_f}, to get the predicted rating for a user-item pair. The model can then be trained for recommendation in a completely supervised manner, using a differentiable cost function such as \eqref{eq:cost_mf}. 
Note that a key difference between this approach and the existing approaches which use item content \cite{wang2011collaborative, wang2015collaborative}, apart from sensitivity to word order, is that we do not define an unsupervised objective (like likelihood of observing bag-of-words under a topic model) for extracting a text representation. 
However, our model can benefit from unsupervised data through pre-training of word embeddings \cite{mikolov2013distributed} or pre-training of RNN parameters using language models \cite{dai2015semi} (our experiments use embeddings). 

\subsubsection{Gated Recurrent Units (GRUs)}

Traditional RNN architectures suffer from the problem of vanishing and exploding gradients \cite{bengio1994learning}, rendering optimization difficult and prohibiting them from learning long-term dependencies. 
There have been several modifications to the RNN proposed to remedy this problem, of which the most popular are \textit{long short-term memory units} (LSTMs) \cite{hochreiter1997long} and the more recent \textit{gated recurrent units} (GRUs) \cite{cho2014learning}. We use GRUs, which are simpler than LSTM, have fewer parameters, and give competitive performance to LSTMs \cite{chung2014empirical, jozefowicz2015empirical}.

The GRU hidden vector output at step $t$, $h_t$, for the input sequence $X_j = (w_1, \ldots, w_t, \ldots, w_{n_j})$ is given by:
\begin{align}
	\begin{bmatrix}
	f_t \\
	o_t \\
	\end{bmatrix}
	&= \sigma\left( \theta^1
	\begin{bmatrix}
	 \ee_{w_t} \\
	 h_{t-1} \\
	\end{bmatrix}
	 +
	 b
	 \right) \label{eq:gates} \\
	 c_t &= \tanh(\theta^2_w \ee_{w_t}  + f_t \odot \theta^2_h h_{t-1} + b_c) \label{eq:cell}\\
	 h_t &= (1 - o_t) \odot h_{t-1} + o_t \odot c_t \label{eq:hidden}
\end{align}
where $\theta^1 \in \RR^{2K_h \times (K_w + K_h)}, \theta^2_w \in \RR^{K_h \times K_w}, \theta^2_h \in \RR^{K_h \times K_h}$  and $b, b_c \in \RR^{K_h}$ are parameters of the GRU with $K_w$ the dimension of input word embeddings and $K_h$ the number of hidden units in the RNN. $\odot$ denotes element-wise product. Intuitively, $f_t$ \eqref{eq:gates} acts as a `forget' (or `reset') gate that decides what parts of the previous hidden state to consider or ignore at the current step, $c_t$ \eqref{eq:cell} computes a candidate state for the current time step using the parts of the previous hidden state as dictated by $f_t$, and $o_t$ \eqref{eq:gates} acts as the output (or update) gate which decides what parts of the previous memory to change to the new candidate memory \eqref{eq:hidden}. All forget and update operations are differentiable to allow learning through backpropagation.

The final architecture, shown in Figure \ref{fig:model}, consists of two stacked layers of RNN with GRU hidden units. 
We use a bi-directional RNN \cite{schuster1997bidirectional} at the first layer and feed the concatenation of the forward and backward hidden states as the input to the second layer.
The output of the hidden states of the second layer is pooled to obtain the item content representation $g(X_j)$. In our experiments, mean pooling performs best. Models that use the final RNN state take much longer to optimize. Following~\eqref{eq:embed_avg_shift}, the final item representation is obtained by combining the RNN representation with an item-specific embedding $v_j$.
We now describe the multi-task learning setup.

\subsection{Multi-Task Learning}
\label{sec:mtl2}
The encoder $f(\cdot)$ can be used as a generic feature extractor for items. Therefore, we can employ the multi-task learning approach of Section~\ref{sec:back-mtl}. The tags associated with papers can be considered as a (coarse) summary or topics of the items and thus forcing the encoder to be predictive of the tags will provide a useful inductive bias.   Consider again the toy example of Figure \ref{fig:model}. Observing the tag ``RNN'' but not ``LDA'' on the paper, even though the term LDA is present in the text, will force the network to pay attention to the sequence of words ``not LDA'' in order to explain the tags. 

We define the probability of observing tag $l$ on item $j$ as: $P(t_{jl} = 1) = p_{jl} = \sigma(f(X_j)^T\te_l)$, where $\te_{l}$ is an embedding for tag $l$. 
The cost for predicting the tags is taken as the sum of the weighted binary log likelihood of each tag:
\begin{align*}
C_T(\pmb{\theta}) = \frac{1}{|T|} \sum_{j} \sum_{l} \{t_{jl} \log p_{jl} + c_{jl}' (1-t_{jl}) \log (1-p_{jl}) \}
\end{align*}
where $c_{jl}'$ down-weights the cost for predicting the unobserved tags.
The final cost is $C(\pmb{\theta}) = \lambda C_R(\pmb{\theta}) + (1-\lambda) C_T(\pmb{\theta})$ with $C_R$ defined in \eqref{eq:cost_mf}, and $\lambda$ is a hyperparameter.

It is worth noting the differences between our approach and \citet{almahairi2015learning},
who use language modeling on the text as an unsupervised multi-task objective with the item latent factors as the shared parameters.
\citet{almahairi2015learning} found that the increased flexibility offered by the RNN makes it too strong a regularizer leading to worse performance than simpler bag-of-words models. In contrast, our RNN is trained fully supervised, which forces the item representations to be discriminative  for recommendation and tag prediction.  Furthermore, by using the text as an input to $g(\cdot)$ at test time, rather than just for train-time regularization, we can alleviate the cold-start problem. 


\section{Experiments}
\begin{table*}[t!]
	\centering
	\caption{\% Recall@50 for all the methods (higher is better).}
	\label{tab:main}
	\begin{tabular}{|c||c|c|c||c|c|c|}
		\hline
		& \multicolumn{3}{c||}{Citeulike-a} & \multicolumn{3}{c|}{Citeulike-t} \\ \cline{2-7}
		& Warm Start & Cold Start & Tag Prediction & Warm Start & Cold Start & Tag Prediction \\ \hline
		GRU-MTL   & \textbf{38.33} & \textbf{49.76} & \textbf{60.52} & 45.60 & \textbf{51.22} & \textbf{62.32} \\ \hline
		GRU	   	  & 36.87 & 46.16 & ----- & 42.59 & 47.59 & ----- \\ \hline
		CTR-MTL   & 35.51 & 39.87 & 48.95 & \textbf{46.82} & 34.98 & 46.66 \\ \hline
		CTR  	  & 31.10 & 39.00 & ----- & 40.44 & 33.74 & ----- \\ \hline
		Embed-MTL & 36.64 & 41.71 & 60.36 & 43.02 & 38.16 & 62.29 \\ \hline
		Embed     & 33.95 & 38.53 & ----- & 37.98 & 35.85 & ----- \\ \hline
		
	\end{tabular}
	
\end{table*}

\begin{figure*}[t!]
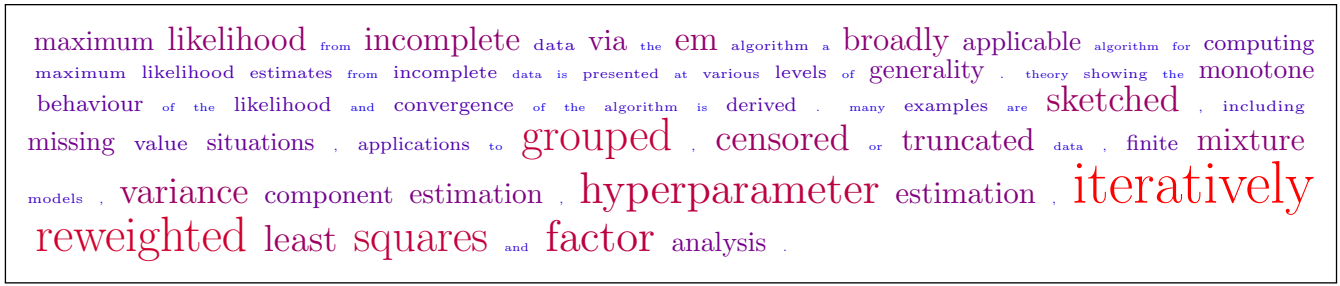

\begin{framed}

{\textcolor[rgb]{0.46, 0.00, 0.54}{\fontsize{0.37cm}{1cm} \selectfont maximum}} {\textcolor[rgb]{0.57, 0.00, 0.43}{\fontsize{0.45cm}{1cm} \selectfont likelihood}} {\textcolor[rgb]{0.14, 0.00, 0.86}{\fontsize{0.15cm}{1cm} \selectfont from}} {\textcolor[rgb]{0.56, 0.00, 0.44}{\fontsize{0.45cm}{1cm} \selectfont incomplete}} {\textcolor[rgb]{0.26, 0.00, 0.74}{\fontsize{0.21cm}{1cm} \selectfont data}} {\textcolor[rgb]{0.51, 0.00, 0.49}{\fontsize{0.40cm}{1cm} \selectfont via}} {\textcolor[rgb]{0.10, 0.00, 0.90}{\fontsize{0.15cm}{1cm} \selectfont the}} {\textcolor[rgb]{0.58, 0.00, 0.42}{\fontsize{0.46cm}{1cm} \selectfont em}} {\textcolor[rgb]{0.23, 0.00, 0.77}{\fontsize{0.18cm}{1cm} \selectfont algorithm}} {\textcolor[rgb]{0.12, 0.00, 0.88}{\fontsize{0.15cm}{1cm} \selectfont a}} {\textcolor[rgb]{0.55, 0.00, 0.45}{\fontsize{0.44cm}{1cm} \selectfont broadly}} {\textcolor[rgb]{0.45, 0.00, 0.55}{\fontsize{0.36cm}{1cm} \selectfont applicable}} {\textcolor[rgb]{0.19, 0.00, 0.81}{\fontsize{0.15cm}{1cm} \selectfont algorithm}} {\textcolor[rgb]{0.12, 0.00, 0.88}{\fontsize{0.15cm}{1cm} \selectfont for}} {\textcolor[rgb]{0.37, 0.00, 0.63}{\fontsize{0.30cm}{1cm} \selectfont computing}} {\textcolor[rgb]{0.27, 0.00, 0.73}{\fontsize{0.21cm}{1cm} \selectfont maximum}} {\textcolor[rgb]{0.30, 0.00, 0.70}{\fontsize{0.24cm}{1cm} \selectfont likelihood}} {\textcolor[rgb]{0.27, 0.00, 0.73}{\fontsize{0.22cm}{1cm} \selectfont estimates}} {\textcolor[rgb]{0.07, 0.00, 0.93}{\fontsize{0.15cm}{1cm} \selectfont from}} {\textcolor[rgb]{0.30, 0.00, 0.70}{\fontsize{0.24cm}{1cm} \selectfont incomplete}} {\textcolor[rgb]{0.14, 0.00, 0.86}{\fontsize{0.15cm}{1cm} \selectfont data}} {\textcolor[rgb]{0.05, 0.00, 0.95}{\fontsize{0.15cm}{1cm} \selectfont is}} {\textcolor[rgb]{0.23, 0.00, 0.77}{\fontsize{0.18cm}{1cm} \selectfont presented}} {\textcolor[rgb]{0.11, 0.00, 0.89}{\fontsize{0.15cm}{1cm} \selectfont at}} {\textcolor[rgb]{0.22, 0.00, 0.78}{\fontsize{0.18cm}{1cm} \selectfont various}} {\textcolor[rgb]{0.30, 0.00, 0.70}{\fontsize{0.24cm}{1cm} \selectfont levels}} {\textcolor[rgb]{0.08, 0.00, 0.92}{\fontsize{0.15cm}{1cm} \selectfont of}} {\textcolor[rgb]{0.45, 0.00, 0.55}{\fontsize{0.36cm}{1cm} \selectfont generality}} {\textcolor[rgb]{0.06, 0.00, 0.94}{\fontsize{0.15cm}{1cm} \selectfont .}} {\textcolor[rgb]{0.16, 0.00, 0.84}{\fontsize{0.15cm}{1cm} \selectfont theory}} {\textcolor[rgb]{0.22, 0.00, 0.78}{\fontsize{0.18cm}{1cm} \selectfont showing}} {\textcolor[rgb]{0.06, 0.00, 0.94}{\fontsize{0.15cm}{1cm} \selectfont the}} {\textcolor[rgb]{0.45, 0.00, 0.55}{\fontsize{0.36cm}{1cm} \selectfont monotone}} {\textcolor[rgb]{0.38, 0.00, 0.62}{\fontsize{0.31cm}{1cm} \selectfont behaviour}} {\textcolor[rgb]{0.07, 0.00, 0.93}{\fontsize{0.15cm}{1cm} \selectfont of}} {\textcolor[rgb]{0.07, 0.00, 0.93}{\fontsize{0.15cm}{1cm} \selectfont the}} {\textcolor[rgb]{0.34, 0.00, 0.66}{\fontsize{0.27cm}{1cm} \selectfont likelihood}} {\textcolor[rgb]{0.06, 0.00, 0.94}{\fontsize{0.15cm}{1cm} \selectfont and}} {\textcolor[rgb]{0.33, 0.00, 0.67}{\fontsize{0.27cm}{1cm} \selectfont convergence}} {\textcolor[rgb]{0.08, 0.00, 0.92}{\fontsize{0.15cm}{1cm} \selectfont of}} {\textcolor[rgb]{0.08, 0.00, 0.92}{\fontsize{0.15cm}{1cm} \selectfont the}} {\textcolor[rgb]{0.21, 0.00, 0.79}{\fontsize{0.17cm}{1cm} \selectfont algorithm}} {\textcolor[rgb]{0.07, 0.00, 0.93}{\fontsize{0.15cm}{1cm} \selectfont is}} {\textcolor[rgb]{0.32, 0.00, 0.68}{\fontsize{0.26cm}{1cm} \selectfont derived}} {\textcolor[rgb]{0.09, 0.00, 0.91}{\fontsize{0.15cm}{1cm} \selectfont .}} {\textcolor[rgb]{0.15, 0.00, 0.85}{\fontsize{0.15cm}{1cm} \selectfont many}} {\textcolor[rgb]{0.28, 0.00, 0.72}{\fontsize{0.23cm}{1cm} \selectfont examples}} {\textcolor[rgb]{0.09, 0.00, 0.91}{\fontsize{0.15cm}{1cm} \selectfont are}} {\textcolor[rgb]{0.61, 0.00, 0.39}{\fontsize{0.49cm}{1cm} \selectfont sketched}} {\textcolor[rgb]{0.09, 0.00, 0.91}{\fontsize{0.15cm}{1cm} \selectfont ,}} {\textcolor[rgb]{0.25, 0.00, 0.75}{\fontsize{0.20cm}{1cm} \selectfont including}} {\textcolor[rgb]{0.45, 0.00, 0.55}{\fontsize{0.36cm}{1cm} \selectfont missing}} {\textcolor[rgb]{0.35, 0.00, 0.65}{\fontsize{0.28cm}{1cm} \selectfont value}} {\textcolor[rgb]{0.41, 0.00, 0.59}{\fontsize{0.33cm}{1cm} \selectfont situations}} {\textcolor[rgb]{0.10, 0.00, 0.90}{\fontsize{0.15cm}{1cm} \selectfont ,}} {\textcolor[rgb]{0.31, 0.00, 0.69}{\fontsize{0.25cm}{1cm} \selectfont applications}} {\textcolor[rgb]{0.09, 0.00, 0.91}{\fontsize{0.15cm}{1cm} \selectfont to}} {\textcolor[rgb]{0.77, 0.00, 0.23}{\fontsize{0.62cm}{1cm} \selectfont grouped}} {\textcolor[rgb]{0.09, 0.00, 0.91}{\fontsize{0.15cm}{1cm} \selectfont ,}} {\textcolor[rgb]{0.61, 0.00, 0.39}{\fontsize{0.49cm}{1cm} \selectfont censored}} {\textcolor[rgb]{0.09, 0.00, 0.91}{\fontsize{0.15cm}{1cm} \selectfont or}} {\textcolor[rgb]{0.52, 0.00, 0.48}{\fontsize{0.42cm}{1cm} \selectfont truncated}} {\textcolor[rgb]{0.14, 0.00, 0.86}{\fontsize{0.15cm}{1cm} \selectfont data}} {\textcolor[rgb]{0.07, 0.00, 0.93}{\fontsize{0.15cm}{1cm} \selectfont ,}} {\textcolor[rgb]{0.37, 0.00, 0.63}{\fontsize{0.29cm}{1cm} \selectfont finite}} {\textcolor[rgb]{0.54, 0.00, 0.46}{\fontsize{0.43cm}{1cm} \selectfont mixture}} {\textcolor[rgb]{0.23, 0.00, 0.77}{\fontsize{0.18cm}{1cm} \selectfont models}} {\textcolor[rgb]{0.09, 0.00, 0.91}{\fontsize{0.15cm}{1cm} \selectfont ,}} {\textcolor[rgb]{0.61, 0.00, 0.39}{\fontsize{0.49cm}{1cm} \selectfont variance}} {\textcolor[rgb]{0.44, 0.00, 0.56}{\fontsize{0.35cm}{1cm} \selectfont component}} {\textcolor[rgb]{0.49, 0.00, 0.51}{\fontsize{0.39cm}{1cm} \selectfont estimation}} {\textcolor[rgb]{0.11, 0.00, 0.89}{\fontsize{0.15cm}{1cm} \selectfont ,}} {\textcolor[rgb]{0.74, 0.00, 0.26}{\fontsize{0.59cm}{1cm} \selectfont hyperparameter}} {\textcolor[rgb]{0.51, 0.00, 0.49}{\fontsize{0.41cm}{1cm} \selectfont estimation}} {\textcolor[rgb]{0.12, 0.00, 0.88}{\fontsize{0.15cm}{1cm} \selectfont ,}} {\textcolor[rgb]{1.00, 0.00, 0.00}{\fontsize{0.80cm}{1cm} \selectfont iteratively}} {\textcolor[rgb]{0.82, 0.00, 0.18}{\fontsize{0.66cm}{1cm} \selectfont reweighted}} {\textcolor[rgb]{0.62, 0.00, 0.38}{\fontsize{0.50cm}{1cm} \selectfont least}} {\textcolor[rgb]{0.76, 0.00, 0.24}{\fontsize{0.61cm}{1cm} \selectfont squares}} {\textcolor[rgb]{0.18, 0.00, 0.82}{\fontsize{0.15cm}{1cm} \selectfont and}} {\textcolor[rgb]{0.74, 0.00, 0.26}{\fontsize{0.59cm}{1cm} \selectfont factor}} {\textcolor[rgb]{0.46, 0.00, 0.54}{\fontsize{0.37cm}{1cm} \selectfont analysis}} {\textcolor[rgb]{0.19, 0.00, 0.81}{\fontsize{0.15cm}{1cm} \selectfont .}}

\end{framed}
\caption{Saliency of each word in the abstract of the EM paper~\cite{dempster1977maximum}. Size and color of the words indicate their leverage on the final rating. The model learns that chunks of word phrases are important, such as ``maximum likelihood'' and ``iteratively reweighted least squares'', and ignores punctutations and stop words.}
\label{fig:heatmap}
\end{figure*}

\subsection{Experimental Setup}
\textbf{Datasets:}
We use two datasets made available by \citet{wang2015collaborative} from CiteULike\footnote{http://www.citeulike.org/}. CiteULike is an online platform which allows registered users to create personal libraries by saving papers which are of interest to them.
The datasets consist of the papers in the users' libraries (which are treated as `likes'), user provided tags on the papers, and the title and abstract of the papers.
Similar to \citet{wang2011collaborative}, we remove users with less than 5 ratings (since they cannot be evaluated properly) and removed tags that occur on less than 10 articles.
\textit{Citeulike-a} \cite{wang2011collaborative} consists of 5551 users, 16980 papers and 3629 tags with a total of 
204,987 user-item likes.
\textit{Citeulike-t} \cite{wang2011collaborative} consists of 5219 users, 25975 papers and 4222 tags with a total of 
134,860 user-item likes.
Note \textit{Citeulike-t} is much more sparse (99.90\%) than \textit{Citeulike-a} (99.78\%).

\textbf{Evaluation Methodology:}
Following, \citet{wang2011collaborative}, we test the models on held-out user-article likes under both warm-start and cold-start scenarios.

\textit{Warm-Start:} 
This is the case of in-matrix prediction, where every test item had at least one like in the training data. For each user we do a 5-fold split of papers from their like history.
Papers with less than 5 likes are always kept in the training data, since they cannot be evaluated properly.
After learning, we predict ratings across all active test set items and for each user filter out the items in their training set from the ranked list.

\textit{Cold-Start:} This is the task of predicting user interest in a new paper with no existing likes, based on the text content of the paper.
The set of all papers is split into 5 folds. 
Again, papers with less than 5 likes are always kept in training set.
For each fold, we
remove all likes on the papers in that fold forming
the test-set and keep the other folds as training-set.
We fit the models on the training set items for each fold and form predictive per-user ranking of items in the test set.

\textit{Evaluation Metric:}
Accuracy of recommendation from implicit feedback is often measured by recall. Precision is not reasonable since the zero ratings may mean that a user either does not like the article or does not know of it. Thus, we use Recall@M \cite{wang2011collaborative} and average the per-user metric:
\begin{align*}
	\mbox{Recall@}M = \frac{\mbox{number of articles user liked in top } M}{\mbox{total number of articles user liked}}
\end{align*}

\subsubsection{Methods}
We compare the proposed methods with \textit{CTR}, which models item content using topic modeling. The approach put forth by CTR~\cite{wang2011collaborative} cannot perform tag-prediction and thus, for a fair comparison, we modify CTR to do tag prediction. This can be viewed as a probabilistic version of collective matrix factorization \cite{singh2008relational}.
Deriving an alternating least squares inference algorithm along the line of \cite{wang2011collaborative} is not possible for a sigmoid loss. Thus, for CTR, we formulate tag prediction using a weighted squared loss instead. Learning this model is a straightforward extension of CTR: rather than performing alternating updates on two blocks of parameters, we rotate among three. We call this \textit{CTR-MTL}.
The word embedding-based model with order-insensitive document encoder (section \ref{sec:order_insensitive}) is \textit{Embed}, and the RNN-based model (section \ref{sec:order_sensitive}) is \textit{GRU}. The corresponding models trained with multi-task learning are~\textit{Embed-MTL} and \textit{GRU-MTL}.

\begin{figure*}[t!]
\center
\includegraphics[scale=0.21]{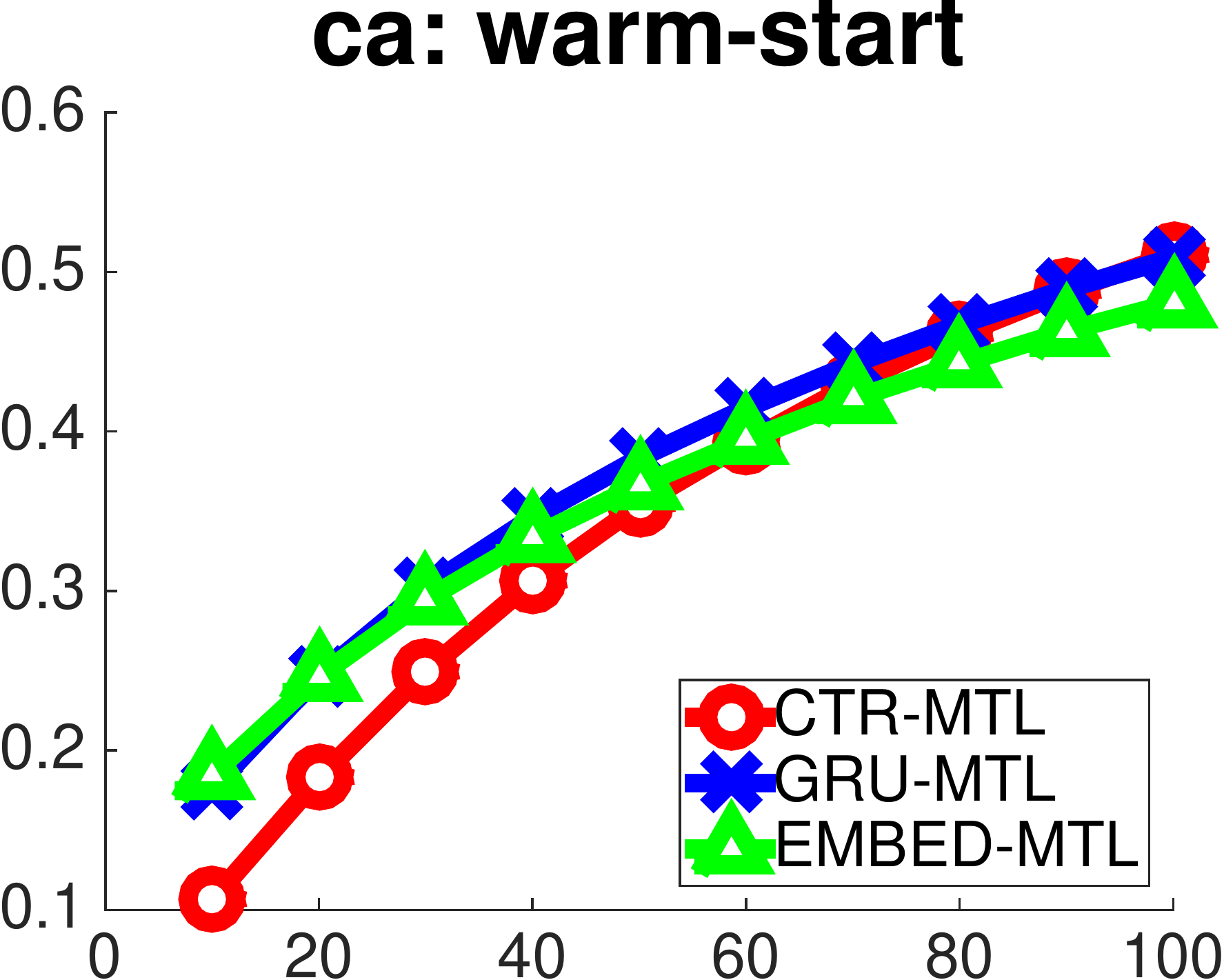}
\hspace{0.2em}
\includegraphics[scale=0.21]{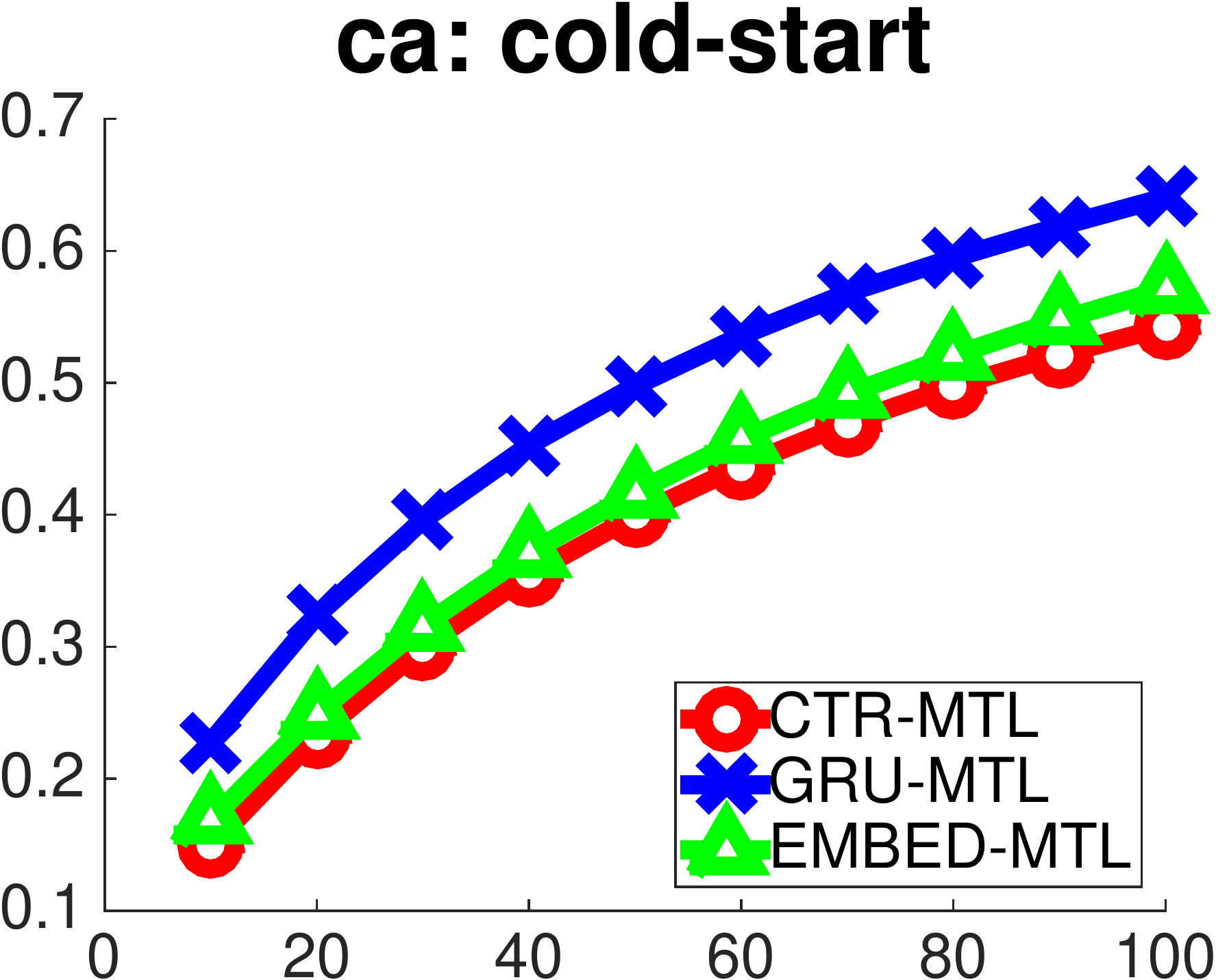}
\hspace{0.2em}
\includegraphics[scale=0.21]{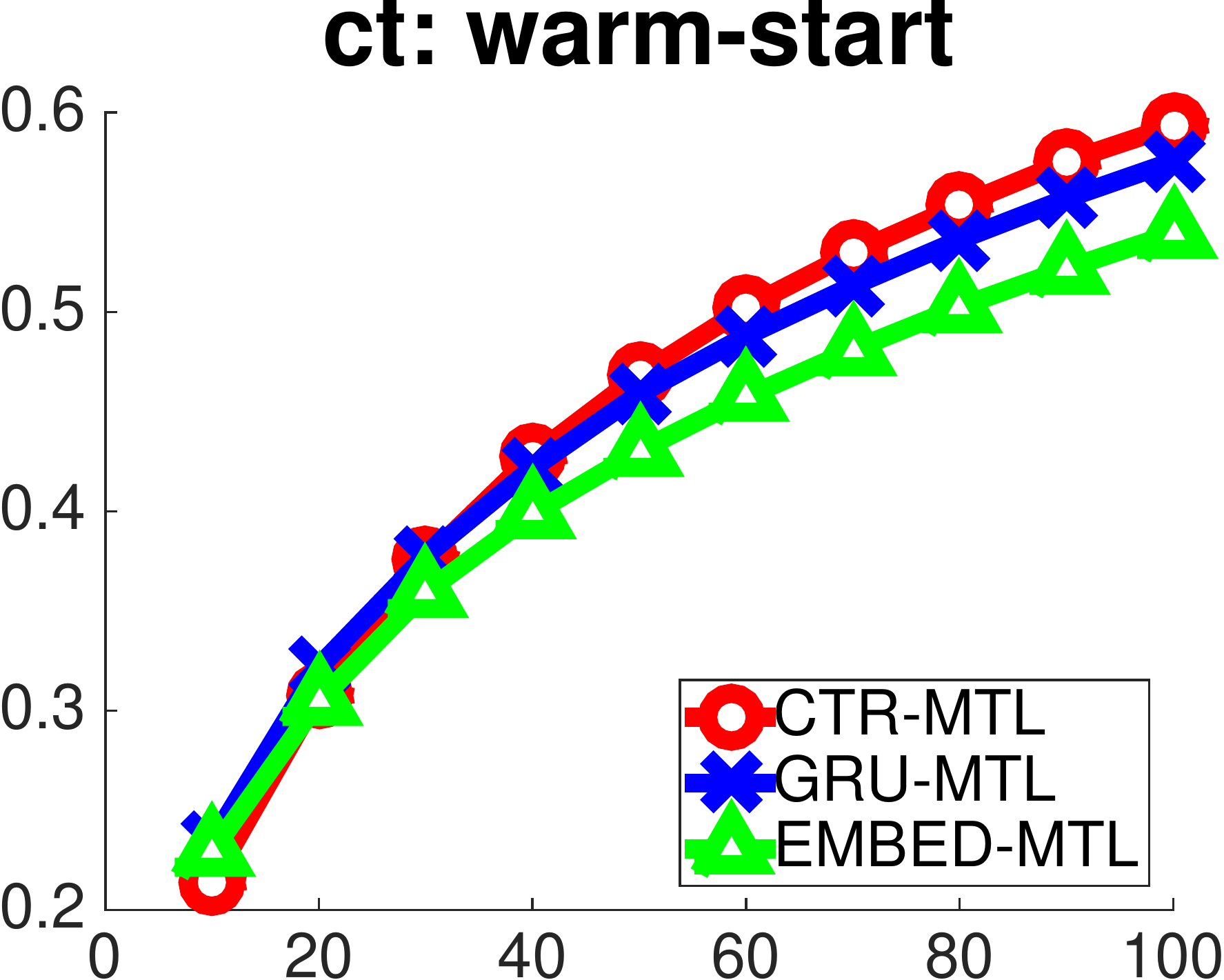}
\hspace{0.2em}
\includegraphics[scale=0.21]{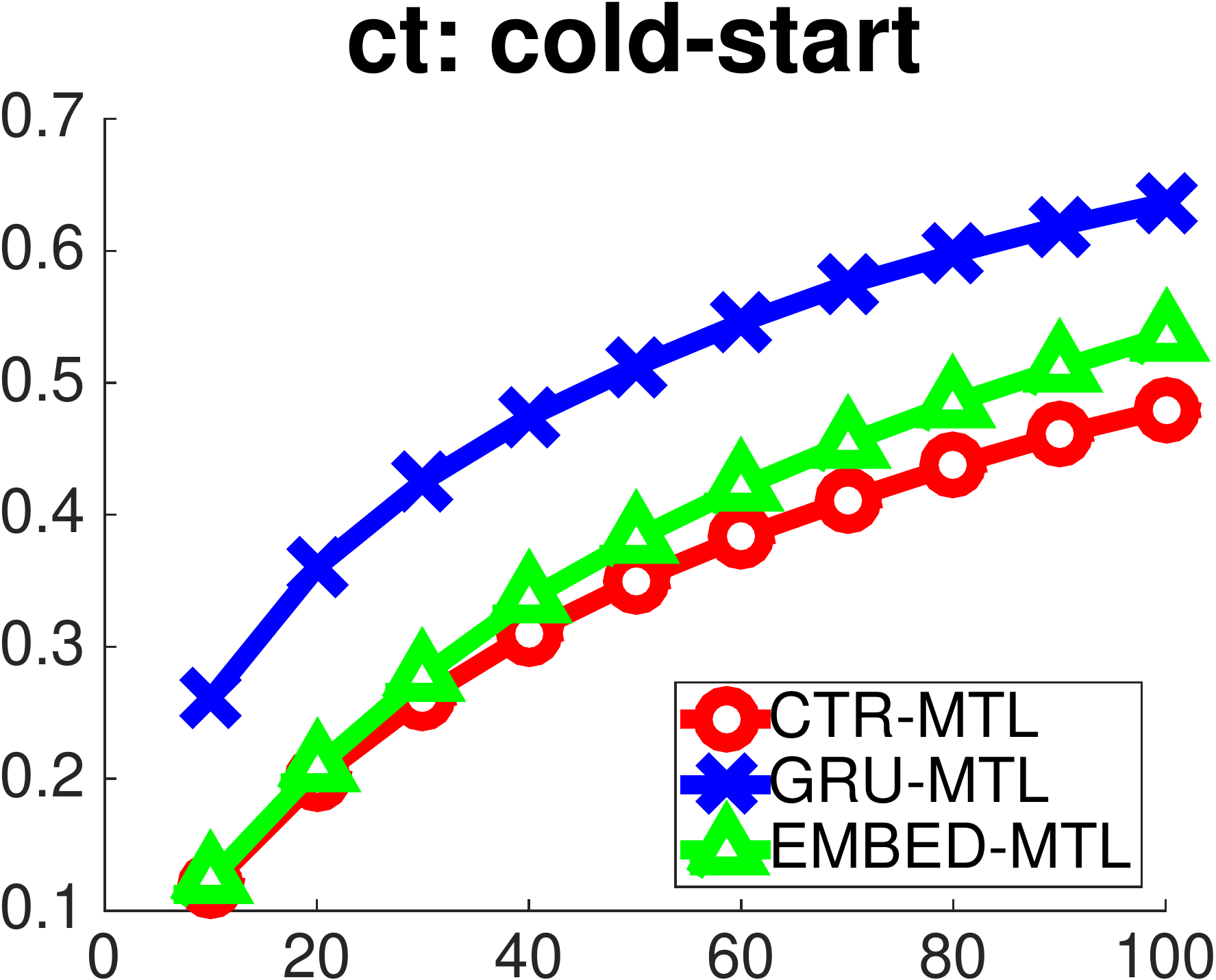}
	\caption{Recall@M for the models trained with multi-task learning. $x$-axis is the value of $M \in [100]$}.
	\label{fig:curves}
\end{figure*}

\subsubsection{Implementation Details}
For CTR, we follow~\citet{wang2011collaborative} for setting hyperparameters. We use latent factor dimension $K=200$, regularization parameters $\lambda_u=0.01,\; \lambda_v=100$ and cost weights $a=1,\; b=0.01$. The same parameters  gave good results for CTR-MTL. CTR and CTR-MTL are trained using the EM algorithm, which updates the latent factors using alternating least squares on full data \cite{wang2011collaborative, hu2008collaborative}. CTR is sensitive to good pre-processing of the text, which is common in topic modeling~\cite{wang2011collaborative}. We use the provided pre-processed text for CTR, which was obtained by removing stop-words and choosing top words based on tf-idf. We initialized CTR with the output of a topic model trained only on the text.
We used the CTR code provided by the authors.

For the \textit{Embed} and \textit{GRU} models, we used word embeddings of dimension $K_w = K = 200$, in order to be consistent with CTR. For \textit{GRU} models, the first layer of the RNN has hidden state dimension $K_{h1} = 400$ and the second layer (the output layer) has hidden state dimension $K_{h2} = 200$. We pre-trained the word embeddings using CBOW \cite{mikolov2013distributed} on a corpus of 440,756 ACM abstracts (including the Citeulike abstracts). 
Dropout is used at every layer of the network. The probabilities of dropping a dimension are 0.1, 0.5 and 0.3 at the embedding layer, the output of the first layer and the output of the second layer, respectively. We also regularize the user embeddings with weight $0.01$.
We do very mild preprocessing of the text. We replace numbers with a <NUM> token and all words which have a total frequency of less than 5 by <UNK>.
Note that we don't remove stop words or frequent words. This leaves a vocabulary of 21,129 words for \textit{Citeulike-a} and 24,697 words for \textit{Citeulike-t}.

The models are optimized via stochastic gradient descent, where mini-batch randomly samples a subset of $B$ users and for each user we sample one positive and one negative example. 
We set the weights $c_{ij}$ in \eqref{eq:cost_mf} to $c_{ij} = 1 + \alpha \log (1 + \frac{|R_i|}{\epsilon})$, where $|R_i|$ is the number of items liked by user $i$, with $\alpha=10,\; \epsilon=1e-8$. Unlike \citet{wang2011collaborative} we do not weight the cost function differently for positive and negative samples. Since the total number of negative examples is much larger than the positive examples for each user, stochastically sampling only one negative per positive example implicitly down-weights the negatives.
We used a mini-batch size of $B=512$ users and used Adam \cite{kingma2014adam} for optimization. We run the models for a maximum of 20k mini-batch updates and use early-stopping based on recall on a validation set from the training examples.

\subsection{Quantitative Results}
Table \ref{tab:main} summarizes Recall@50 for all the models, on the two CiteULike datasets, for both warm-start and cold-start.
Figure~\ref{fig:curves}, further shows the variation of Recall@M for different values of $M$ for the multi-task learning models.

\textbf{Cold-Start}:
Recall that for cold-start recommendation, the item-specific embeddings $\ve_j$ in~\eqref{eq:embed_avg_shift} are identically equal to zero, and thus the items' representations depend solely on their text content. We first note the performance of the models without multi-task learning.
The GRU model is better than the best score of either the CTR model or the Embed model by \textbf{18.36\%} (relative improvement) on CiteUlike-a and by \textbf{32.74\%} on CiteULike-t.
This significant gain demonstrates that the GRU model is much better at representing the content of the items.
Improvements are higher on the CitULike-t dataset because it is much more sparse, and so models which can utilize content appropriately give better recommendations. CTR and Embed models perform competitively with each other.

Next, observe that multi-task learning uniformly improves performance for all models. The GRU model's recall improves by 7.8\% on Citulike-a and by 7.6\% on Citeulike-t. This leads to an overall improvement of \textbf{19.30\%} on Citeulike-a and \textbf{34.22\%} on Citeulike-t, over best of the baselines.
Comparatively, improvement for CTR is smaller. This is expected since the Bayesian topic model provides strong regularization for the model parameters. Contrary to this, Embed models also benefits a lot by MTL (up to 8.2\%). This is expected since unlike CTR, all the $K_w\times V$ parameters in the Embed model are free parameters which are trained directly for recommendation, and thus MTL provides necessary regularization.

\textbf{Warm-Start}:
Collaborative filtering methods based on matrix factorization \cite{koren2009matrix} often perform as well as hybrid methods in the warm-start scenario, due to the flexibility of the item-specific embeddings $\ve_j$ in~\eqref{eq:embed_avg_shift} \cite{wang2011collaborative, wang2014improving}.
Consider again the models trained without MTL.
GRU model performs better than either the CTR or the Embed model, with relative improvement of \textbf{8.5\%} on CiteULike-a and \textbf{5.3\%} on CiteULike-t, over the best of the two models.
Multi-task learning again improves performance for all the models.
Improvements are particularly significant for CTR-MTL over CTR (up to 15.8\%).
Since the tags associated with test items were observed during training, they provide a strong inductive bias leading to improved performance.
Interestingly, the GRU-MTL model performs slightly better than the CTR-MTL model on one dataset and slightly worse on the other.
The first and third plots in Figure~\ref{fig:curves} demonstrate that the GRU-MTL performs slightly better than the CTR-MTL for smaller $M$, i.e. more relevant articles are ranked toward the top.
To quantify this, we evaluate average reciprocal Hit-Rank@10 \cite{bansal2015content}.
Given a list of $M$ ranked articles for
user $i$ , let $c_1, c_2, \ldots, c_h$ denote the
ranks of $h$ articles in $[M]$ which the user actually liked.
HR is then defined as $\sum_{t=1}^{h}\frac{1}{c_t}$
and tests whether
top ranked articles are correct.
GRU-MTL gives HR@10 of \textbf{0.098} and CTR-MTL gives HR@10 to be 0.077,  which confirms that the top of the list for GRU-MTL contains more relevant recommendations. 

\textbf{Tag Prediction:}
Although the focus of the models is recommendation, we evaluate the performance of the multi-task models on tag prediction. We again use Recall@50 (defined per article) and evaluate in the cold-start scenario, where there are no tags present for the test article. The GRU and Embed models perform similarly. CTR-MTL is significantly worse, which could be due to our use of the squared loss for training or because hyperparameters were selected for recommendation performance, not tag prediction.

\subsection{Interpreting Prediction Decisions}
\label{sec:heatmap}
 We employ a simple, easy-to-implement tool for analyzing RNN predictions, based on~\citet{denil2014extraction} and \citet{li2015visualizing}. We produce a heatmap where every input word is associated with its leverage on the output prediction. Suppose that we recommended item $j$ to user $i$. In other words, suppose that $\hat{r}_{ij}$ is large. Let $E_j = (e_{j,1}, e_{j,2}, \ldots, e_{j,n_j})$ be the sequence of word embeddings for item $j$.  
Since $f(\cdot)$ is encoded as a neural network, $\frac{d \hat{r}_{ij}}{d e_{j,t}}$ can be obtained by backpropagation. To produce the heatmap's value for word $t$, we convert $\frac{d \hat{r}_{ij}}{d e_{j,t}}$ into a scalar. This is not possible by backpropagation, as $\frac{d \hat{r}_{ij}}{d x_{j,t}}$ is not well-defined, since $x_{j,t}$ is a discrete index. Instead we compute $\lVert \frac{d \hat{r}_{ij}}{d e_{j,t}} \rVert$. An application is in Figure \ref{fig:heatmap}.

\section{Conclusion \& Future Work}
We employ deep recurrent neural networks to provide vector representations for the text content associated with items in collaborative filtering. This generic text-to-vector mapping is useful because it can be trained directly with gradient descent and provides opportunities to perform multi-task learning. For scientific paper recommendation, the RNN and multi-task learning both provide complementary performance improvements. We encourage further use of the technique in a variety of application domains. 
In future work, we would like to apply deep architectures to users' data and to explore additional objectives for multi-task learning that employ multiple modalities of inputs, such as movies' images and text descriptions. 

\section{Acknowledgment}
This work was supported in part by the Center for Intelligent Information Retrieval, in part by The Allen Institute for Artificial Intelligence, in part by NSF grant \#CNS-0958392, in part by the National Science Foundation (NSF) grant number DMR-1534431, and in part by DARPA under agreement number FA8750-13-2-0020. The U.S. Government is authorized to reproduce and distribute reprints for Governmental purposes notwithstanding any copyright notation thereon. Any opinions, findings and conclusions or recommendations expressed in this material are those of the authors and do not necessarily reflect those of the sponsor.

%
\bibliographystyle{unsrtnat}

\bibliography{refs}  

\begin{thebibliography}{56}
\providecommand{\natexlab}[1]{#1}
\providecommand{\url}[1]{\texttt{#1}}
\expandafter\ifx\csname urlstyle\endcsname\relax
  \providecommand{\doi}[1]{doi: #1}\else
  \providecommand{\doi}{doi: \begingroup \urlstyle{rm}\Url}\fi

\bibitem[Guy et~al.(2010)Guy, Zwerdling, Ronen, Carmel, and
  Uziel]{guy2010social}
Ido Guy, Naama Zwerdling, Inbal Ronen, David Carmel, and Erel Uziel.
\newblock Social media recommendation based on people and tags.
\newblock In \emph{SIGIR}, 2010.

\bibitem[Phelan et~al.(2009)Phelan, McCarthy, and Smyth]{phelan2009using}
Owen Phelan, Kevin McCarthy, and Barry Smyth.
\newblock Using twitter to recommend real-time topical news.
\newblock In \emph{RecSys}, 2009.

\bibitem[Bansal et~al.(2015)Bansal, Das, and Bhattacharyya]{bansal2015content}
Trapit Bansal, Mrinal Das, and Chiranjib Bhattacharyya.
\newblock Content driven user profiling for comment-worthy recommendations of
  news and blog articles.
\newblock In \emph{RecSys}, 2015.

\bibitem[McAuley and Leskovec(2013)]{mcauley2013hidden}
Julian McAuley and Jure Leskovec.
\newblock Hidden factors and hidden topics: understanding rating dimensions
  with review text.
\newblock In \emph{RecSys}, 2013.

\bibitem[Wang and Blei(2011)]{wang2011collaborative}
Chong Wang and David~M Blei.
\newblock Collaborative topic modeling for recommending scientific articles.
\newblock In \emph{SIGKDD}, 2011.

\bibitem[Koren et~al.(2009)Koren, Bell, and Volinsky]{koren2009matrix}
Yehuda Koren, Robert Bell, and Chris Volinsky.
\newblock Matrix factorization techniques for recommender systems.
\newblock \emph{Computer}, \penalty0 (8):\penalty0 30--37, 2009.

\bibitem[Mnih and Salakhutdinov(2007)]{mnih2007probabilistic}
Andriy Mnih and Ruslan Salakhutdinov.
\newblock Probabilistic matrix factorization.
\newblock In \emph{NIPS}, 2007.

\bibitem[Balabanovi{\'c} and Shoham(1997)]{balabanovic1997fab}
Marko Balabanovi{\'c} and Yoav Shoham.
\newblock Fab: content-based, collaborative recommendation.
\newblock \emph{Communications of the ACM}, 40\penalty0 (3):\penalty0 66--72,
  1997.

\bibitem[Mooney and Roy(2000)]{mooney2000content}
Raymond~J Mooney and Loriene Roy.
\newblock Content-based book recommending using learning for text
  categorization.
\newblock In \emph{ACM conference on Digital libraries}, 2000.

\bibitem[Basu et~al.(1998)Basu, Hirsh, Cohen, et~al.]{basu1998recommendation}
Chumki Basu, Haym Hirsh, William Cohen, et~al.
\newblock Recommendation as classification: Using social and content-based
  information in recommendation.
\newblock In \emph{AAAI}, 1998.

\bibitem[Schein et~al.(2002)Schein, Popescul, Ungar, and
  Pennock]{schein2002methods}
Andrew~I Schein, Alexandrin Popescul, Lyle~H Ungar, and David~M Pennock.
\newblock Methods and metrics for cold-start recommendations.
\newblock In \emph{SIGIR}, 2002.

\bibitem[Basilico and Hofmann(2004)]{basilico2004unifying}
Justin Basilico and Thomas Hofmann.
\newblock Unifying collaborative and content-based filtering.
\newblock In \emph{ICML}, 2004.

\bibitem[Wang et~al.(2015)Wang, Wang, and Yeung]{wang2015collaborative}
Hao Wang, Naiyan Wang, and Dit-Yan Yeung.
\newblock Collaborative deep learning for recommender systems.
\newblock In \emph{SIGKDD}, 2015.

\bibitem[Melville et~al.(2002)Melville, Mooney, and
  Nagarajan]{melville2002content}
Prem Melville, Raymond~J Mooney, and Ramadass Nagarajan.
\newblock Content-boosted collaborative filtering for improved recommendations.
\newblock In \emph{AAAI}, 2002.

\bibitem[Gopalan et~al.(2014)Gopalan, Charlin, and Blei]{gopalan2014content}
Prem~K Gopalan, Laurent Charlin, and David Blei.
\newblock Content-based recommendations with poisson factorization.
\newblock In \emph{NIPS}, 2014.

\bibitem[Agarwal and Chen(2009)]{agarwal2009regression}
Deepak Agarwal and Bee-Chung Chen.
\newblock Regression-based latent factor models.
\newblock In \emph{SIGKDD}, 2009.

\bibitem[Wallach(2006)]{wallach2006topic}
Hanna~M Wallach.
\newblock Topic modeling: beyond bag-of-words.
\newblock In \emph{ICML}, 2006.

\bibitem[Werbos(1990)]{werbos1990backpropagation}
Paul~J Werbos.
\newblock Backpropagation through time: what it does and how to do it.
\newblock \emph{Proceedings of the IEEE}, 78\penalty0 (10):\penalty0
  1550--1560, 1990.

\bibitem[Mikolov et~al.(2010)Mikolov, Karafi{\'a}t, Burget, Cernock{\`y}, and
  Khudanpur]{mikolov2010recurrent}
Tomas Mikolov, Martin Karafi{\'a}t, Lukas Burget, Jan Cernock{\`y}, and Sanjeev
  Khudanpur.
\newblock Recurrent neural network based language model.
\newblock \emph{INTERSPEECH}, 2010.

\bibitem[Cho et~al.(2014)Cho, van Merrienboer, Gulcehre, Bougares, Schwenk, and
  Bengio]{cho2014learning}
Kyunghyun Cho, Bart van Merrienboer, Caglar Gulcehre, Fethi Bougares, Holger
  Schwenk, and Yoshua Bengio.
\newblock Learning phrase representations using rnn encoder-decoder for
  statistical machine translation.
\newblock In \emph{EMNLP}, 2014.

\bibitem[Dai and Le(2015)]{dai2015semi}
Andrew~M Dai and Quoc~V Le.
\newblock Semi-supervised sequence learning.
\newblock In \emph{NIPS}, pages 3061--3069, 2015.

\bibitem[Bell and Koren(2007)]{bell2007lessons}
Robert~M Bell and Yehuda Koren.
\newblock Lessons from the netflix prize challenge.
\newblock \emph{SIGKDD Explorations Newsletter}, 9\penalty0 (2):\penalty0
  75--79, 2007.

\bibitem[Ling et~al.(2014)Ling, Lyu, and King]{ling2014ratings}
Guang Ling, Michael~R Lyu, and Irwin King.
\newblock Ratings meet reviews, a combined approach to recommend.
\newblock In \emph{RecSys}, 2014.

\bibitem[Almahairi et~al.(2015)Almahairi, Kastner, Cho, and
  Courville]{almahairi2015learning}
Amjad Almahairi, Kyle Kastner, Kyunghyun Cho, and Aaron Courville.
\newblock Learning distributed representations from reviews for collaborative
  filtering.
\newblock In \emph{RecSys}, 2015.

\bibitem[Weston et~al.(2011)Weston, Bengio, and Usunier]{weston2011wsabie}
Jason Weston, Samy Bengio, and Nicolas Usunier.
\newblock Wsabie: Scaling up to large vocabulary image annotation.
\newblock In \emph{IJCAI}, 2011.

\bibitem[Hu et~al.(2008)Hu, Koren, and Volinsky]{hu2008collaborative}
Yifan Hu, Yehuda Koren, and Chris Volinsky.
\newblock Collaborative filtering for implicit feedback datasets.
\newblock In \emph{ICDM}, 2008.

\bibitem[Rendle et~al.(2009)Rendle, Freudenthaler, Gantner, and
  Schmidt-Thieme]{rendle2009bpr}
Steffen Rendle, Christoph Freudenthaler, Zeno Gantner, and Lars Schmidt-Thieme.
\newblock Bpr: Bayesian personalized ranking from implicit feedback.
\newblock In \emph{UAI}, 2009.

\bibitem[Shi et~al.(2014)Shi, Larson, and Hanjalic]{shi2014collaborative}
Yue Shi, Martha Larson, and Alan Hanjalic.
\newblock Collaborative filtering beyond the user-item matrix: A survey of the
  state of the art and future challenges.
\newblock \emph{ACM Computing Surveys}, 47\penalty0 (1):\penalty0 3, 2014.

\bibitem[Rendle(2010)]{rendle2010factorization}
Steffen Rendle.
\newblock Factorization machines.
\newblock In \emph{ICDM}, 2010.

\bibitem[Gantner et~al.(2010)Gantner, Drumond, Freudenthaler, Rendle, and
  Schmidt-Thieme]{gantner2010learning}
Zeno Gantner, Lucas Drumond, Christoph Freudenthaler, Steffen Rendle, and Lars
  Schmidt-Thieme.
\newblock Learning attribute-to-feature mappings for cold-start
  recommendations.
\newblock In \emph{ICDM}, 2010.

\bibitem[Caruana(1997)]{caruana1997multitask}
Rich Caruana.
\newblock Multitask learning.
\newblock \emph{Machine learning}, 28\penalty0 (1):\penalty0 41--75, 1997.

\bibitem[Singh and Gordon(2008)]{singh2008relational}
Ajit~P Singh and Geoffrey~J Gordon.
\newblock Relational learning via collective matrix factorization.
\newblock In \emph{SIGKDD}, 2008.

\bibitem[Ma et~al.(2008)Ma, Yang, Lyu, and King]{ma2008sorec}
Hao Ma, Haixuan Yang, Michael~R Lyu, and Irwin King.
\newblock Sorec: social recommendation using probabilistic matrix
  factorization.
\newblock In \emph{CIKM}, 2008.

\bibitem[Krestel et~al.(2009)Krestel, Fankhauser, and Nejdl]{krestel2009latent}
Ralf Krestel, Peter Fankhauser, and Wolfgang Nejdl.
\newblock Latent dirichlet allocation for tag recommendation.
\newblock In \emph{RecSys}, 2009.

\bibitem[Ian~Goodfellow and Courville(2016)]{Goodfellow-et-al-2016-Book}
Yoshua~Bengio Ian~Goodfellow and Aaron Courville.
\newblock Deep learning.
\newblock Book in prep. for MIT Press, 2016.

\bibitem[Salakhutdinov et~al.(2007)Salakhutdinov, Mnih, and
  Hinton]{salakhutdinov2007restricted}
Ruslan Salakhutdinov, Andriy Mnih, and Geoffrey Hinton.
\newblock Restricted boltzmann machines for collaborative filtering.
\newblock In \emph{ICML}, 2007.

\bibitem[Sedhain et~al.(2015)Sedhain, Menon, Sanner, and
  Xie]{sedhain2015autorec}
Suvash Sedhain, Aditya~Krishna Menon, Scott Sanner, and Lexing Xie.
\newblock Autorec: Autoencoders meet collaborative filtering.
\newblock In \emph{WWW}, 2015.

\bibitem[Wu et~al.(2016)Wu, DuBois, Zheng, and Ester]{wu2016cdae}
Yao Wu, Christopher DuBois, Alice~X. Zheng, and Martin Ester.
\newblock Collaborative denoising auto-encoders for top-n recommender systems.
\newblock In \emph{WSDM}, 2016.

\bibitem[Elkahky et~al.(2015)Elkahky, Song, and He]{elkahky2015multi}
Ali~Mamdouh Elkahky, Yang Song, and Xiaodong He.
\newblock A multi-view deep learning approach for cross domain user modeling in
  recommendation systems.
\newblock In \emph{WWW}, 2015.

\bibitem[Dziugaite and Roy(2015)]{dziugaite2015neural}
Gintare~Karolina Dziugaite and Daniel~M Roy.
\newblock Neural network matrix factorization.
\newblock \emph{arXiv preprint arXiv:1511.06443}, 2015.

\bibitem[Van~den Oord et~al.(2013)Van~den Oord, Dieleman, and
  Schrauwen]{van2013deep}
Aaron Van~den Oord, Sander Dieleman, and Benjamin Schrauwen.
\newblock Deep content-based music recommendation.
\newblock In \emph{NIPS}, 2013.

\bibitem[Wang and Wang(2014)]{wang2014improving}
Xinxi Wang and Ye~Wang.
\newblock Improving content-based and hybrid music recommendation using deep
  learning.
\newblock In \emph{International Conference on Multimedia}, 2014.

\bibitem[Weston et~al.(2014)Weston, Chopra, and Adams]{weston2014tagspace}
Jason Weston, Sumit Chopra, and Keith Adams.
\newblock \# tagspace: Semantic embeddings from hashtags.
\newblock 2014.

\bibitem[He and McAuley(2016)]{HeMcA16}
R.~He and J.~McAuley.
\newblock {VBPR:} visual bayesian personalized ranking from implicit feedback.
\newblock In \emph{AAAI}, 2016.

\bibitem[Mikolov et~al.(2013)Mikolov, Sutskever, Chen, Corrado, and
  Dean]{mikolov2013distributed}
Tomas Mikolov, Ilya Sutskever, Kai Chen, Greg~S Corrado, and Jeff Dean.
\newblock Distributed representations of words and phrases and their
  compositionality.
\newblock In \emph{NIPS}, pages 3111--3119, 2013.

\bibitem[Collobert et~al.(2011)Collobert, Weston, Bottou, Karlen, Kavukcuoglu,
  and Kuksa]{collobert2011natural}
Ronan Collobert, Jason Weston, L{\'e}on Bottou, Michael Karlen, Koray
  Kavukcuoglu, and Pavel Kuksa.
\newblock Natural language processing (almost) from scratch.
\newblock \emph{JMLR}, 12:\penalty0 2493--2537, 2011.

\bibitem[Sutskever et~al.(2014)Sutskever, Vinyals, and
  Le]{sutskever2014sequence}
Ilya Sutskever, Oriol Vinyals, and Quoc~V Le.
\newblock Sequence to sequence learning with neural networks.
\newblock In \emph{NIPS}, pages 3104--3112, 2014.

\bibitem[Bengio et~al.(1994)Bengio, Simard, and Frasconi]{bengio1994learning}
Yoshua Bengio, Patrice Simard, and Paolo Frasconi.
\newblock Learning long-term dependencies with gradient descent is difficult.
\newblock \emph{Neural Networks}, 5\penalty0 (2):\penalty0 157--166, 1994.

\bibitem[Hochreiter and Schmidhuber(1997)]{hochreiter1997long}
Sepp Hochreiter and J{\"u}rgen Schmidhuber.
\newblock Long short-term memory.
\newblock \emph{Neural computation}, 9\penalty0 (8):\penalty0 1735--1780, 1997.

\bibitem[Chung et~al.(2014)Chung, Gulcehre, Cho, and
  Bengio]{chung2014empirical}
Junyoung Chung, Caglar Gulcehre, KyungHyun Cho, and Yoshua Bengio.
\newblock Empirical evaluation of gated recurrent neural networks on sequence
  modeling.
\newblock \emph{arXiv preprint arXiv:1412.3555}, 2014.

\bibitem[Jozefowicz et~al.(2015)Jozefowicz, Zaremba, and
  Sutskever]{jozefowicz2015empirical}
Rafal Jozefowicz, Wojciech Zaremba, and Ilya Sutskever.
\newblock An empirical exploration of recurrent network architectures.
\newblock In \emph{ICML}, 2015.

\bibitem[Schuster and Paliwal(1997)]{schuster1997bidirectional}
Mike Schuster and Kuldip~K Paliwal.
\newblock Bidirectional recurrent neural networks.
\newblock \emph{Signal Processing}, 45\penalty0 (11):\penalty0 2673--2681,
  1997.

\bibitem[Dempster et~al.(1977)Dempster, Laird, and Rubin]{dempster1977maximum}
Arthur~P Dempster, Nan~M Laird, and Donald~B Rubin.
\newblock Maximum likelihood from incomplete data via the em algorithm.
\newblock \emph{Journal of the royal statistical society.}, pages 1--38, 1977.

\bibitem[Kingma and Ba(2014)]{kingma2014adam}
Diederik Kingma and Jimmy Ba.
\newblock Adam: A method for stochastic optimization.
\newblock \emph{arXiv preprint arXiv:1412.6980}, 2014.

\bibitem[Denil et~al.(2014)Denil, Demiraj, and de~Freitas]{denil2014extraction}
Misha Denil, Alban Demiraj, and Nando de~Freitas.
\newblock Extraction of salient sentences from labelled documents.
\newblock \emph{arXiv preprint arXiv:1412.6815}, 2014.

\bibitem[Li et~al.(2016)Li, Chen, Hovy, and Jurafsky]{li2015visualizing}
Jiwei Li, Xinlei Chen, Eduard Hovy, and Dan Jurafsky.
\newblock Visualizing and understanding neural models in nlp.
\newblock 2016.

\end{thebibliography}
\end{document}